\documentclass[letterpaper, 10 pt, conference]{ieeeconf}  

\IEEEoverridecommandlockouts                             
\usepackage{graphics}
\usepackage[dvipsnames]{xcolor}
\usepackage{epsfig} 
\usepackage{mathptmx} 
\usepackage{times} 
\usepackage{amsmath} 
\usepackage{amssymb} 
\usepackage{subcaption}
\usepackage{graphicx}
\usepackage{float}
\usepackage{bm}
\usepackage{caption}
\usepackage{mathtools}
\usepackage{stfloats}
\usepackage{mathrsfs}
\usepackage{multirow}
\usepackage{makecell}
\usepackage{vcell}
\usepackage{booktabs}
\usepackage{diagbox}
\usepackage{csquotes}
\usepackage{cite}
\usepackage{tikz}
\usepackage{lipsum}
\usepackage{schemata}
\usepackage{tabularray}
\usepackage{amsmath}
\usepackage{dcolumn} 
\usepackage{graphicx}

\usepackage[hyphens]{url}
\usepackage[hidelinks]{hyperref}
\hypersetup{breaklinks=true}
\usepackage{cite}

\usepackage{algorithm}
\usepackage[noend]{algpseudocode}

\algrenewcommand\algorithmicforall{\textbf{for each}}
\algrenewcommand\algorithmicindent{.8em}

\algdef{SE}[DOWHILE]{Do}{doWhile}{\algorithmicdo}[1]{\algorithmicwhile\ #1}%
\algnewcommand\algorithmicforeach{\textbf{for each}}
\algdef{S}[FOR]{ForEach}[1]{\algorithmicforeach\ #1\ \algorithmicdo}

\usetikzlibrary{decorations.pathreplacing,calc}

\usepackage{hyperref}
\hypersetup{
colorlinks=true,
linkcolor=blue,
filecolor=magenta,
urlcolor=blue,
}

\usepackage[font=footnotesize]{caption}

\title{\LARGE \bf
SMART-LLM: Smart Multi-Agent Robot Task Planning using Large Language Models
}
\author{Shyam Sundar Kannan$^\dagger$, Vishnunandan L. N. Venkatesh$^\dagger$, and Byung-Cheol Min
\thanks{This material is based upon work supported by the National Science Foundation under Grant No. IIS-1846221. The authors are with SMART Lab, Department of Computer and Information Technology, Purdue University, West Lafayette, IN 47907, USA {\tt\small \{kannan9,lvenkate,minb\}@purdue.edu}}%
\thanks{$\dagger$ Equal contribution.}
}

\begin{document}
\maketitle
\thispagestyle{empty}
\pagestyle{empty}

\begin{abstract}
In this work, we introduce SMART-LLM, an innovative framework designed for embodied multi-robot task planning. SMART-LLM: Smart Multi-Agent Robot Task Planning using Large Language Models (LLMs), harnesses the power of LLMs to convert high-level task instructions provided as input into a multi-robot task plan. It accomplishes this by executing a series of stages, including task decomposition, coalition formation, and task allocation, all guided by programmatic LLM prompts within the few-shot prompting paradigm. We create a benchmark dataset designed for validating the multi-robot task planning problem, encompassing four distinct categories of high-level instructions that vary in task complexity. Our evaluation experiments span both simulation and real-world scenarios, demonstrating that the proposed model can achieve promising results for generating multi-robot task plans. The experimental videos, code, and datasets from the work can be found at \href{https://sites.google.com/view/smart-llm/}{https://sites.google.com/view/smart-llm/}.
\end{abstract}


\section{Introduction}
\label{sec:intro}
In recent years, multi-robot systems have gained prominence in various applications, from housekeeping tasks \cite{benavidez2015design} to search and rescue missions \cite{queralta2020collaborative} and warehouse automation \cite{chen2021integrated}. These systems, composed of multiple autonomous robots, can greatly enhance efficiency, scalability, and adaptability in numerous tasks. Typically, these robot arrays exhibit heterogeneity in terms of types and skill levels among individual agents. Consequently, the overall system complexity is heightened, emphasizing the critical importance of skillful task allocation among these agents. Effective allocation of complex tasks among multiple agents involves several crucial steps, including task decomposition, assigning sub-tasks to suitable agents, and ensuring correct task sequencing \cite{rizk2019cooperative}. This proficiency requires access to external knowledge or domain-specific information about the task.

\begin{figure}[t]
    \centering
    \includegraphics[width=0.97\linewidth]{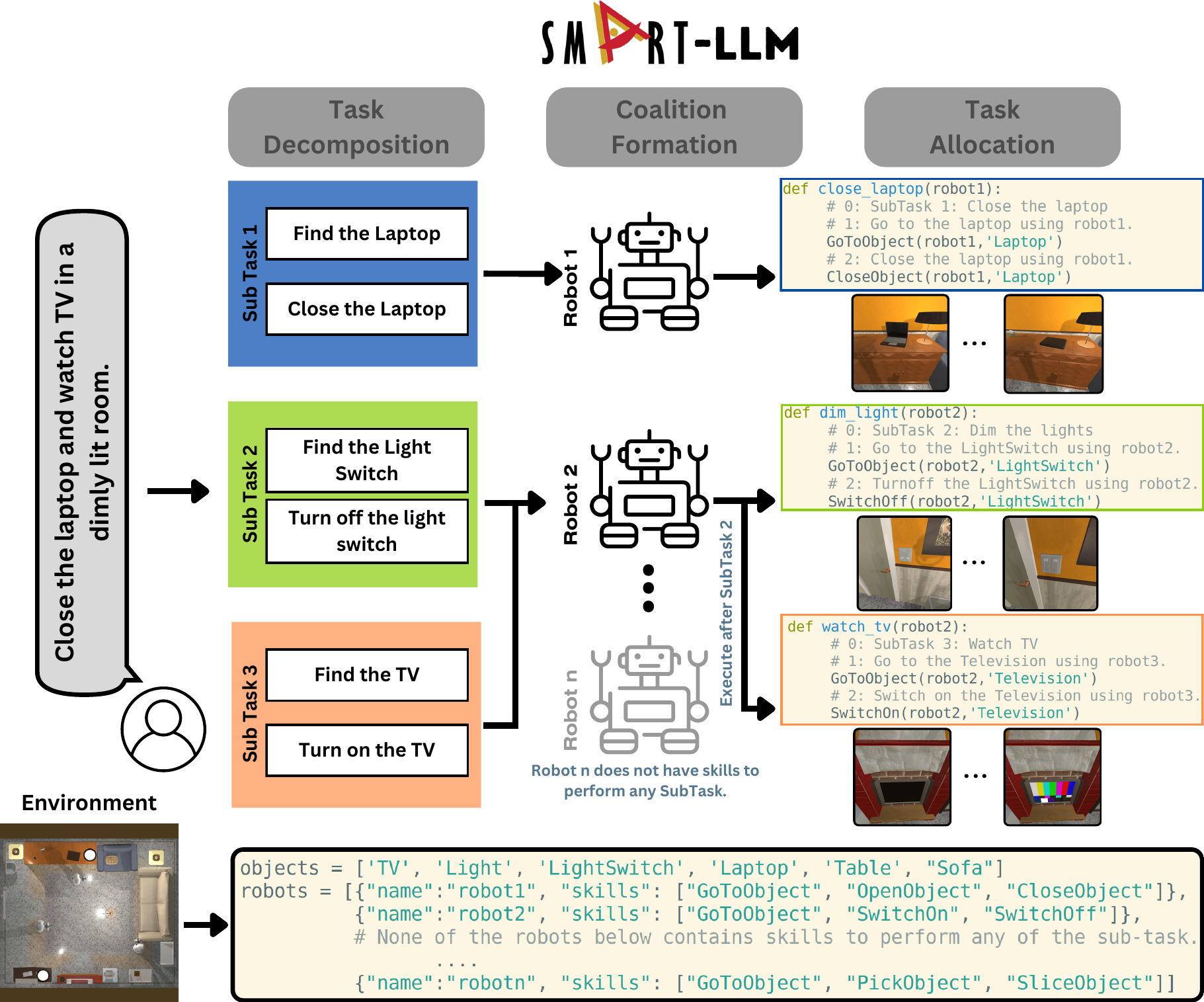}
    \caption{\textbf{An overview of SMART-LLM}: Smart Multi-Agent Robot Task planning using Large Language Models (LLM). Given a high-level instruction, SMART-LLM decomposes the instruction into sub-tasks assigning them to individual robots based on their specific skills and capabilities, and orchestrating their execution in a coherent and logical sequence.}  
    \label{fig:intro_pic}
    \vspace{-4mm}
\end{figure}

Traditional multi-robot task planning often struggles with diverse tasks and complex environments \cite{rizk2019cooperative}, relying on fixed algorithms. 
Relying on fixed algorithms complicates the process of transitioning from one task to another without substantial modifications to the code. These challenges intensify when tasks are described in natural language, as such descriptions can lack precision and completeness.  Take, for instance, the task presented in Fig.~\ref{fig:intro_pic}: \textit{``Closing the laptop and watching TV in a dimly lit room"}. Notably, this task description does not explicitly mention turning off the lights before watching TV. Given the incomplete and ambiguous nature of the instruction, it is crucial to leverage extensive prior knowledge to interpret the task and aid in efficient task planning.

Large language models (LLMs), such as GPT-4 \cite{openai2023gpt4}, GPT-3.5 \cite{brown2020language} and Llama2 \cite{touvron2023llama}, have demonstrated remarkable capabilities in understanding natural language, logical reasoning, and generalization. This presents exciting opportunities for enhancing comprehension and planning in multi-robot systems. In this paper, we introduce SMART-LLM, an innovative mechanism for task assignment to embodied agents using LLMs. SMART-LLM provides LLMs with Python programming scripts that encapsulate intricate robot skills and environmental details, including object information. It also provides practical examples of task decomposition and allocation based on the robot's capabilities and the environment. Leveraging programming language structures, SMART-LLM taps into the vast dataset of internet code snippets and documentation available to LLMs. As illustrated in Fig.~\ref{fig:intro_pic}, when dealing with a complex task, SMART-LLM divides the task into sub-tasks, each related to specific objects or actions. These sub-tasks are then combined and delegated to suitable robots with the necessary skills to perform them.

The main contributions of this work are three-fold:
\begin{itemize}
    \item \textbf{Multi-Robot Task Planning Framework} for integrating task decomposition, coalition formation, and skill-based task assignment, by leveraging LLMs. 

    \item \textbf{Benchmark Dataset:} A benchmark dataset designed for evaluating multi-agent task planning systems, covering a spectrum of tasks, ranging from elemental to complex ones in the AI2-THOR \cite{kolve2017ai2} simulation platform.

    \item \textbf{Implementation and Evaluation} of the framework in both simulated and real-world settings, undergoing thorough testing across a wide array of tasks.
\end{itemize}

\section{Related Works}
\label{sec:rel_work}
\noindent\textbf{Multi-Robot Task Planning.~} Multi-robot task planning is important in robotics, requiring effective coordination among robots. Typically, the process of multi-robot task planning encompasses four distinct phases: task decomposition, coalition formation, task allocation, and task execution \cite{rizk2019cooperative}. Task decomposition entails the subdivision of a given task into manageable sub-tasks. The decomposition methods can either be task-specific \cite{motes2020multi} or necessitate copious amounts of data for generating policies \cite{shiarlis2018taco}. Task-specific decomposition methods cannot be generalized and gathering prior knowledge to decompose a diverse range of tasks can pose a significant challenge. A modern, intuitive strategy employs natural language to articulate tasks and utilizes pre-trained language models equipped with diverse domain knowledge to segment them into sub-tasks and predict their sequential order over time \cite{jansen2020visually, sakaguchi2021proscript}. Similarly, in SMART-LLM, we employ large-language models to deconstruct tasks into robot actions aligned with their skills, facilitating seamless execution by the robot.

In coalition formation and task allocation, efficiently assigning decomposed tasks to multiple agents is crucial for the effective completion of the given task. To this end, a plethora of methodologies have been employed, encompassing negotiation \cite{kong2015negotiation}, auctioning \cite{braquet2021greedy}, consensus-based strategies \cite{zitouni2020distributed} and reinforcement learning \cite{qin2021multi}. 
While these methods exhibit reliability, they are typically tailored and optimized according to specific end goals and applications. This necessitates additional effort when scaling them across different applications and optimizing them under varied constraints. In our approach, we take advantage of the inherent generalizability of LLMs. This enables tasks to scale seamlessly, allowing teams to be assigned in diverse configurations without imposing additional modifications to the constraints within the code.

Degrees of automation, contingent on the number of task-planning steps a method can execute, have been conceptualized \cite{rizk2019cooperative}. Most methods predominantly fall into the first or second level of automation. The first level exclusively automates task execution \cite{barrett2015cooperating, stegagno2013relative}. Meanwhile, the second level automates either task allocation and execution \cite{das2015distributed, mina2020adaptive}; or coalition formation and execution \cite{liu2016coalition}. The third level of automation encompasses coalition, allocation, and execution but does not involve task decomposition \cite{jones2006dynamically, padmanabhan2015coalition}. In a pioneering stride towards the fourth level of automation \cite{liu2022embodied}, a method adeptly manages all four facets of task planning using natural language prompts and Long Short Term Memory (LSTM). Existing methods in the literature often have shortcomings, such as not covering all task planning steps; requiring extensive task-specific demonstration data for model training \cite{liu2022embodied}, which often lacks generalizability when faced with unseen tasks, or being limited to specific tasks. Our method stands out by efficiently performing all four task-planning steps and utilizing LLMs to generalize across various tasks through a few-shot training approach.

\noindent\textbf{LLMs for robotics.~} Large Language Models excel in generalization, commonsense reasoning \cite{brown2020language, madaan2022language}, and are increasingly sought after for inclusion in robotics systems \cite{huang2022language, ahn2022can}. They play a vital role in crafting task plans for robots, making use of few or zero-shot learning methods \cite{brown2020language}. Various techniques for generating these robotic task plans using LLMs have emerged, encompassing value function-based approaches \cite{ahn2022can, lin2023text2motion} and context-driven prompts \cite{singh2023progprompt, chen2023open, wu2023tidybot, huang2023instruct2act}. Moreover, LLMs have found utility in providing feedback and refining task plans to enhance robot performance \cite{vemprala2023chatgpt, huang2022inner, yao2022react}.

While LLMs excel at creating flexible task plans, they face challenges when applied to larger multi-agent teams. In the realm of multi-agent systems, progress has been made in enhancing agent cooperation with the use of LLMs \cite{talebirad2023multi, liu2023bolaa, hong2023metagpt}. These approaches involve equipping individual agents with their own LLMs to improve interactions and boost their collaborative skills. However, these methods prioritize improving multi-agent system efficiency but do not tackle the specific task of creating task plans for multi-robot teams. These plans involve assigning and sequencing tasks for individual robots based on their skills and the environment's condition. Our approach focuses on task decomposition and allocation in a heterogeneous robot team, considering individual robot skills. We achieve multi-robot task planning without the need for separate LLMs per robot. This simplifies planning and provides a unified solution for multi-robot task coordination.

\section{Problem Formulation}
\label{sec:problem_form}
Given a high-level language instruction $I$, the goal of this work is to understand the instruction, compute the necessary steps for task completion, and formulate a task plan that enables its execution. Tasks are executed in a manner that maximizes the utilization of the available robots, by performing tasks in parallel when feasible. These tasks are performed in an environment $E$ that encapsulates numerous entities and objects. We assume that the given instruction $I$ can be successfully executed in the environment $E$. 

To execute the task, we have a set of $N$ heterogeneous embodied robot agents $\mathbb{R} = \{R^1, R^2, ..., R^N\}$. Let $\Delta$ be the set of all skills or actions that an agent may be capable of performing. In this work, we assume that robot skills, $\Delta$, are either pre-implemented in the system or that there are available API calls to execute these skills. Each of the agents possesses a diverse set of skills, $\mathbb{S} = \{S^1, S^2, ..., S^N\}$ that they can perform, each subject to specific constraints. Here, $S^n$ represents the list of skills of robot $R^n$, and $S^n \subseteq \Delta$, for $n = 1,2, ... N$. For instance, for the robot skill \texttt{PickUpObject}, there may be constraints on the maximum mass that a robot can pick. 

Now, the instruction $I$ can be decomposed into a temporarily ordered set of $K$ sub-tasks, $\mathbb{T} = \{T^{1}_{t_1}, T^{1}_{t_2}, ..., T^{K}_{t_j}\}$, based on the robot skills, $\Delta$, and the environment $E$, where ${t_j}$ denotes the temporal order of a sub-task and $j$ $\le$ $K$. It is worth noting that some of the sub-tasks can be executed in parallel, having the same temporal precedence. Let $T_S^k$ be the list of skills needed by a robot to complete a sub-task, $T_{t_j}^k$, where $T_S^k \subseteq \Delta$ and $T_{t_j}^k \in \mathbb{T}$. Based on $T_S^k$, the sub-task can be allocated to a robot $R$ with skills $S$ if $T_S^k \subseteq S$, where $R \in \mathbb{R}$ and $S \in \mathbb{S}$. In cases where no single robot satisfies this constraint, a team of two or more robots is required to perform the sub-task. In such scenarios, we form a team of $Q$ robots, $\mathbb{A} = \{A^1, A^2, ..., A^Q\}$, each possessing skills $\mathbb{S}_A = \{S_A^1, S_A^2, ..., S_A^Q\}$, such that $T_S^k \subseteq \bigcup\mathbb{S}_A$.

\begin{figure*}[t]
\centering
\includegraphics[width=0.98\linewidth]{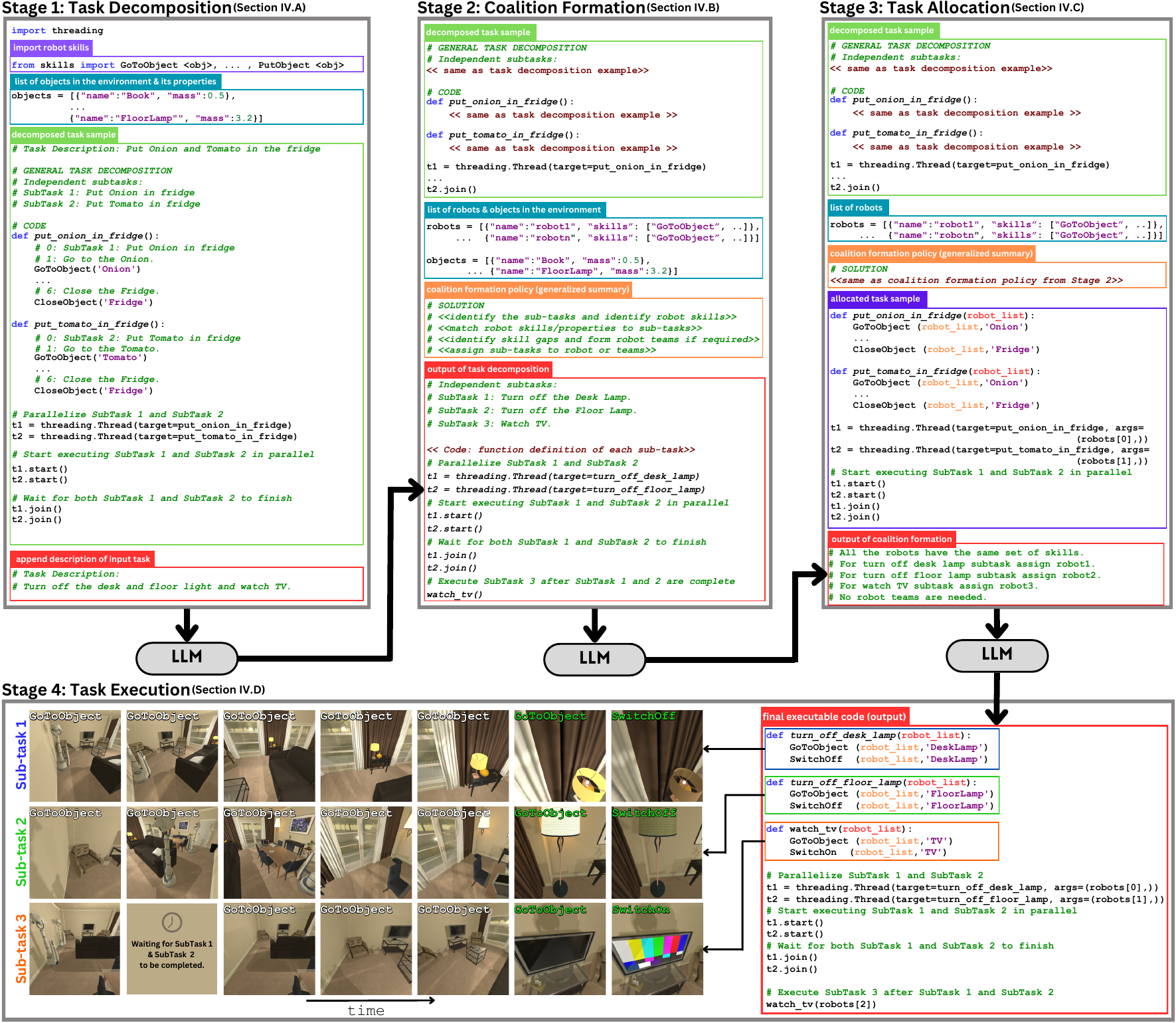}
\caption{\textbf{System overview:} SMART-LLM consists of four key stages: i) Task Decomposition: a prompt consisting of robot skills, objects, and task decomposition samples is combined with the input instruction. This is then fed to the LLM model to decompose the input task; ii) Coalition Formation: a prompt consisting of a list of robots, objects available in the environment, sample decomposed task examples along with corresponding coalition policy describing the formation of robot teams for those tasks, and decomposed task plan for the input task from the previous stage, is given to the LLM, to generate a coalition policy for the input task; iii) Task Allocation: a prompt consisting of sample decomposed tasks, their coalition policy and allocated task plans based on the coalition policy is given to the LLM, along with coalition policy generated for the input task. The LLM then outputs an allocated task plan based on this information; and iv) Task Execution: based on the allocated code generated, the robot executes the tasks. “...” is used for brevity. } 
\label{fig:architecture}
\vspace{-3mm}
\end{figure*}

\section{Methodology}
\label{sec:methodology}
The proposed approach utilizes LLMs to perform \textit{Task Decomposition}, \textit{Coalition Formation}, and \textit{Task Allocation} within the context of multi-robot task planning. Our approach employs Pythonic prompts to guide the LLM in generating code for task decomposition and allocation. We opt for Pythonic prompts over natural language prompts because they facilitate the generation of executable code directly from the LLMs. Moreover, Pythonic prompts adhere to a structured syntax, enhancing the LLM's comprehension of the prompts \cite{wu2023reasoning}. 

We provide concise prompt samples with line-by-line comments and block comments giving task summaries for each step, aiding the LLM in understanding and producing code effectively. The prompts were structured to mimic typical code, complete with comments to delineate sample tasks. Details regarding robot skills and object properties were encoded as Python dictionaries, providing a concise representation that the LLM could readily comprehend \cite{wu2023reasoning} and also help reduce the token size. The comments were meticulously structured, incorporating detailed instructions on task execution and allocation requirements, enabling the LLM to comprehend and replicate the process for new tasks.
\subsection{Stage 1: Task Decomposition}
In this stage, we decompose the given instruction, $I$, into a set of independent sub-tasks, $\mathbb{T}$, along with a sequence of actions for performing each sub-task. To decompose a task, we provide information about the environment, $E$ (including objects and other entities present in the environment), and a list of primitive skills, $\Delta$, that robots can perform. This information about the environment and the robot's skills is utilized to decompose the task such that it can be performed in that environment using the skills possessed by the robots. 

Following the initial few-shot LLM prompting, we provide the LLM with various pieces of information: details about the robot's skills, information about the environment, several examples of sample tasks, and corresponding Python code-based decomposed plans. The LLM takes all this information along with the input task, $I$, that needs to be decomposed and generates the sub-tasks, $\mathbb{T}$. In the Stage 1 block of Fig. \ref{fig:architecture} corresponding to task decomposition, the \textcolor{Orchid}{purple} box corresponds to the list of robot skills, $\Delta$; the \textcolor{cyan}{blue} box corresponds to details about the environment, $E$; \textcolor{LimeGreen}{green} box corresponds to the decomposed task samples given as part of the prompt; and \textcolor{OrangeRed}{red} box corresponds to the given instruction $I$. The \textcolor{OrangeRed}{red} box in the Stage 2 block of Fig. \ref{fig:architecture} is the output from the LLM, corresponding to the sub-tasks, $\mathbb{T}$.

\subsection{Stage 2: Coalition Formation}
Coalition formation is used to form robot teams to perform each of the sub-tasks computed through task decomposition. In task decomposition, the primary task is broken down into sub-tasks, $\mathbb{T}$ based on common sense and the various entities present in the environment, $E$. However, this initial breakdown does not take into account the specific skills of individual robots, $S^n$, or their capabilities to perform each sub-task. Therefore, in this stage, we prompt the LLM to analyze the list of skills needed to perform each sub-task, $T_S^k$, and the skills of individual robots, $S^n$ to identify the suitable robot(s) for each sub-task. To achieve this, we prompt the LLM with samples of decomposed tasks and corresponding coalition formation policies that describe how available robots can be assigned to the sub-tasks. 

The coalition policy consists of statements regarding whether robots possess all the necessary skills to perform a sub-task and how any skill gaps in a single robot's ability to perform a sub-task can be addressed by involving additional robots. 
The samples we include encompass various cases:
\begin{itemize}
    \item In scenarios, where a single robot possesses all the required skills to perform a sub-task, leading to a one-to-one assignment of robots to tasks. 

    \item Instances where no single robot possesses all the skills needed for a sub-task, resulting in multiple robots collaborating on the same task.

    \item Cases where a robot possesses the necessary skills for a sub-task but is constrained by certain limitations (for example, a robot with a maximum weight limit for a pick-up task). In such cases, additional robots are employed to overcome these constraints.
\end{itemize}

By presenting these samples along with the decomposed task, $\mathbb{T}$, and a list of available robots $\mathbb{R}$ and their skills $\mathbb{S}$, the LLM generates a new coalition formation policy that outlines how the given robots can be assigned to perform the input task. The Stage 2 block of Fig. \ref{fig:architecture} corresponding to coalition formation, the \textcolor{LimeGreen}{green} box represents the sample decomposed tasks given as part of the prompt; the \textcolor{cyan}{blue} box shows the available robots $\mathbb{R}$ and their skills $\mathbb{S}$ along with details about the environment $E$; the \textcolor{YellowOrange}{orange} box delineates a general summary of the coalition policy, whereas in the experiments we utilize actual coalition policy for the sample decomposed tasks; and the \textcolor{OrangeRed}{red} box is the decomposed task for which a coalition policy needs to be generated. The \textcolor{OrangeRed}{red} box in the Stage 3 block of Fig. \ref{fig:architecture} is the output from the LLM, corresponding to coalition formation policy for the sub-tasks, $\mathbb{T}$ and the instruction $I$.

\subsection{Stage 3: Task Allocation}
Task allocation involves the precise assignment of either a specific robot or a team of robots to individual sub-tasks, guided by the coalition formation policy established in the preceding phase. Similar to the previous stages, a prompt consisting of decomposed task samples, coalition formation policies, and allocated plans for those tasks is constructed. By incorporating the decomposed sub-tasks, $\mathbb{T}$, and the previously generated coalition formation policy for the given input task, $I$, we instruct the LLM to distribute robots to each sub-task according to the coalitions and produce executable code. Depending on the coalition policy, a sub-task may be allocated to either a single robot or a group of robots. 

The Stage 3 block in Fig. \ref{fig:architecture} shows sample decomposed plans (\textcolor{LimeGreen}{green} box), the list of available robots and their skills (\textcolor{cyan}{blue} box), their coalition policies (\textcolor{YellowOrange}{orange} box), and their allocated plans (\textcolor{Fuchsia}{violet} box) used as part of the prompt, along with the coalition policy for input task (\textcolor{OrangeRed}{red} box), to generate the final executable code in the Stage 4 block (\textcolor{OrangeRed}{red} box).

\subsection{Stage 4: Task Execution}
The LLM generates task plans for multi-robot teams through task allocation, which are then executed by an interpreter with either a virtual or physical team of robots. These plans are executed by making API calls to the robots' low-level skills, ensuring the efficient execution of the tasks. As shown in Stage 4 of Fig. \ref{fig:architecture}, the allocated task plan (\textcolor{OrangeRed}{red} box) for the example task $I$ $=$ \textit{``turn off the desk and floor light and watch TV"} is executed by a team of three robots in a certain temporal order. In this stage, the figure also displays the sequence of robot views as they perform the task along with captions indicating the ongoing task step. Captions marked in \textcolor{green}{green} correspond to specific actions completed by the robot. 


\section{Experiments}
\label{sec:experiments}

\subsection{Benchmark Dataset}
To evaluate the performance of SMART-LLM and facilitate a quantitative comparison with other baseline methods, we created a benchmark dataset tailored for the evaluation of natural language-based task planning in multi-robot scenarios. This dataset originates from environments and actions within AI2-THOR \cite{kolve2017ai2}, a deterministic simulation platform for typical household activities. The dataset encompasses 36 high-level instructions that articulate tasks and corresponding AI2-THOR floor plans, providing the spatial context for task execution. Given the multi-robot facet of our dataset, we include information on the number of robots available to perform a task and a comprehensive list of their respective skills. The number of available robots for each task ranges from 1 to 4, with varying individual skills, allowing for scalability evaluation of task planning methods.

In the dataset, we also include the final ground truth states for the tasks, capturing the definitive states of relevant objects and their conditions within the environment after task completion. This ground truth delineates a set of symbolic goal conditions crucial for achieving task success. It includes details such as the object's position in the environment and its conditions like \texttt{heated}, \texttt{cooked}, \texttt{sliced}, or \texttt{washed} after the task is correctly executed. In addition to the final ground truth states, we provide data on the number of transitions in robot utilization during task execution. Transitions occur when one group of robots completes their sub-tasks, allowing another group to take over. This quantifies the utilization of the multi-robot system. If tasks are not appropriately parallelized during experiments and robots are not fully utilized, sub-tasks may be performed sequentially rather than concurrently, resulting in more transitions in robot utilization compared to ground truth utilization.



To evaluate the performance of our proposed method across diverse task complexities, our dataset comprises four task categories:
\begin{itemize}
    \item \textbf{Elemental Tasks} are designed for a single robot. In these scenarios, a single robot is assumed to possess all the necessary skills and abilities, eliminating the need for coordination with multiple robots.

    \item \textbf{Simple Tasks} involve multiple objects and can be decomposed into sequential or parallel sub-tasks but not both concurrently. Again, all the robots possess all the necessary skills. 
    
    \item \textbf{Compound Tasks} are similar to Simple Tasks, with flexibility in execution strategies (sequential, parallel, or hybrid). However, the robots are heterogeneous, possessing specialized skills and properties, allowing individual robots to handle sub-tasks that match their skills and properties.
    
    \item \textbf{Complex Tasks} are intended for heterogeneous robot teams and resemble Compound Tasks in their characteristics like task decomposition, multi-robot engagement, and the presence of multiple objects. Unlike Compound Tasks, individual robots cannot independently perform sub-tasks due to limitations in their skills or properties, necessitating strategic team assignments to leverage their combined capabilities for effective task completion.
\end{itemize}

The dataset comprises 6 tasks categorized as elemental tasks, 8 tasks as simple tasks, 14 tasks as compound tasks, and 8 tasks as complex tasks.


\begin{table*}[t]
\centering
\large
\setlength\extrarowheight{-3pt}
\caption{Evaluation of SMART-LLM and baselines in the AI2-THOR simulator for different categories of tasks in the benchmark dataset. } 
\label{tab:results}
\resizebox{1.0\textwidth}{!}{
\begin{tabular}{ccccccccccccccccccccccccc}
\hline
\multirow{4}{*}{\textbf{Methods}}                                     &  & \multicolumn{5}{c}{\multirow{2}{*}{ \textbf{Elemental}}}                                                               &  & \multicolumn{5}{c}{\multirow{2}{*}{\textbf{Simple}}}                                                                  &  & \multicolumn{5}{c}{\multirow{2}{*}{\textbf{Compound}}}                                                                &  & \multicolumn{5}{c}{\multirow{2}{*}{\textbf{Complex}}}                                                                  \\
                                                             &  & \multicolumn{5}{c}{}                                                                                          &  & \multicolumn{5}{c}{}                                                                                          &  & \multicolumn{5}{c}{}                                                                                          &  & \multicolumn{5}{c}{}                                                                                           \\ 
\cline{3-7}\cline{9-13}\cline{15-19}\cline{21-25}
                                                             &  & \multirow{2}{*}{\textbf{SR}} & \multirow{2}{*}{\textbf{TCR}} & \multirow{2}{*}{\textbf{GCR}} & \multirow{2}{*}{\textbf{RU}} & \multirow{2}{*}{\textbf{Exe}} &  & \multirow{2}{*}{\textbf{SR}} & \multirow{2}{*}{\textbf{TCR}} & \multirow{2}{*}{\textbf{GCR}} & \multirow{2}{*}{\textbf{RU}} & \multirow{2}{*}{\textbf{Exe}} &  & \multirow{2}{*}{\textbf{SR}} & \multirow{2}{*}{\textbf{TCR}} & \multirow{2}{*}{\textbf{GCR}} & \multirow{2}{*}{\textbf{RU}} & \multirow{2}{*}{\textbf{Exe}} &  & \multirow{2}{*}{\textbf{SR}} & \multirow{2}{*}{\textbf{TCR}} & \multirow{2}{*}{\textbf{GCR}} & \multirow{2}{*}{\textbf{RU}} & \multirow{2}{*}{\textbf{Exe}}  \\
                                                             &  &                     &                      &                      &                     &                     &  &                     &                      &                      &                     &                     &  &                     &                      &                      &                     &                     &  &                     &                      &                      &                     &                      \\ 
\hline
                                                             &  &                     &                      &                      &                     &                     &  &                     &                      &                      &                     &                     &  &                     &                      &                      &                     &                     &  &                     &                      &                      &                     &                      \\
\begin{tabular}[c]{@{}c@{}}{SMART-LLM~}{(GPT-4)}\end{tabular}   &  & \textbf{1.00}                   & \textbf{1.00}           & \textbf{1.00}                    & \textbf{1.00}                   & \textbf{1.00}                   &  & 0.62                   & \textbf{1.00}                    & \textbf{1.00}                    & 0.62                   & \textbf{1.00}                  &  &\textbf{ 0.69}                  & \textbf{0.76}                   & \textbf{0.85}                   & \textbf{0.92}                  & \textbf{1.00}                  &  & \textbf{0.71}                  & \textbf{0.85}                   & \textbf{0.92}                   & \textbf{1.00}                  & \textbf{0.97}                   \\
                                                             &  &                     &                      &                      &                     &                     &  &                     &                      &                      &                     &                     &  &                     &                      &                      &                     &                     &  &                     &                      &                      &                     &                      \\
\begin{tabular}[c]{@{}c@{}}{SMART-LLM~}{(GPT-3.5)}\end{tabular} &  & 0.83                   & 0.83                    & 0.83                    & \textbf{1.00}                   & 0.91                   &  & 0.62                   & 0.87                    & 0.93                    & 0.62                   & 0.95                  &  & 0.42                  & 0.50                  & 0.61                   & 0.71                  & 0.85                  &  & 0.14                  & 0.28                   & 0.35                   & 0.85                  & 0.62                   \\
                                                             &  &                     &                      &                      &                     &                     &  &                     &                      &                      &                     &                     &  &                     &                      &                      &                     &                     &  &                     &                      &                      &                     &                      \\
\begin{tabular}[c]{@{}c@{}}{SMART-LLM~}{(Llama2)}\end{tabular} &  & \textbf{1.00}                   & \textbf{1.00}                    & \textbf{1.00}                    & \textbf{1.00}                   & \textbf{1.00}                   &  & 0.75                   & 0.87                    & 0.93           & 0.87             & \textbf{1.00}         &  & 0.64         & 0.69           & 0.80              & 0.87               & 0.90                  &  & 0.63        & 0.71             & 0.83           & 0.90          & 0.89                   \\
                                                             &  &                     &                      &                      &                     &                     &  &                     &                      &                      &                     &                     &  &                     &                      &                      &                     &                     &  &                     &                      &                      &                     &                      \\       
\begin{tabular}[c]{@{}c@{}}{SMART-LLM~}{(Claude3)}\end{tabular} &  & \textbf{1.00}                   & \textbf{1.00}                    & \textbf{1.00}                    & \textbf{1.00}                   & \textbf{1.00}                   &  & \textbf{0.87}    & \textbf{1.00}                    & \textbf{1.00}                    & \textbf{0.93}                   & \textbf{1.00}         &  & \textbf{0.69}     & \textbf{0.76}      & 0.81       & 0.87                  & \textbf{1.00}                           &  & \textbf{0.71}        & 0.71             & 0.87           & 0.97          & 0.92                   \\
                                                             &  &                     &                      &                      &                     &                     &  &                     &                      &                      &                     &                     &  &                     &                      &                      &                     &                     &  &                     &                      &                      &                     &                      \\                                                             
\begin{tabular}[c]{@{}c@{}}{Decomp~(ours)}{ + Rand}\end{tabular} &  & \textbf{1.00}                   & \textbf{1.00}                    & \textbf{1.00}                    & \textbf{1.00}                   & \textbf{1.00}                   &  & 0.37                   & 0.62                    & 0.62                    & 0.37                   & 0.60                  &  & 0.08                  & 0.16                   & 0.25                   & 0.41                  & 0.37                  &  & 0.00                  & 0.00                   & 0.15                   & 0.85                  & 0.38                   \\
                                                             &  &                     &                      &                      &                     &                     &  &                     &                      &                      &                     &                     &  &                     &                      &                      &                     &                     &  &                     &                      &                      &                     &                      \\
\begin{tabular}[c]{@{}c@{}}{Decomp~(ours)}{ + Rule\cite{gunn2015dynamic}}\end{tabular} &  & \textbf{1.00}                   & \textbf{1.00}                    & \textbf{1.00}                    & \textbf{1.00}                   & \textbf{1.00}                   &  & 0.62                   & \textbf{1.00}                    & \textbf{1.00}                    & 0.62                   & \textbf{1.00}                  &  & 0.57                  & 0.57                   & 0.65                   & 0.81                  & 0.74                  &  & 0.14                  & 0.14                   & 0.35                  & 0.85                  & 0.54                   \\
                                                             &  &                     &                      &                      &                     &                     &  &                     &                      &                      &                     &                     &  &                     &                      &                      &                     &                     &  &                     &                      &                      &                     &                      \\
\hline
\end{tabular}}
\vspace{-1mm}
\end{table*}

\subsection{Simulation Experiments}
Our method's validation takes place within the AI2-THOR simulated environment, where we employ our benchmark dataset for rigorous evaluation and comparative analysis against baseline approaches. Our experimental setup encompasses a varied set of example prompts, including 5 Pythonic plan examples for task decomposition, 3 for coalition formation, and 4 for task allocation. These example prompts cover tasks that can be parallelized using threading, tasks that can only be executed sequentially, and tasks that involve both parallel and sequential execution. This diverse range of examples is strategically tailored to mirror the inherent complexities present in distinct phases of multi-robot task planning. 

It is worth noting that the example prompts were distinct from the tasks in the dataset and were based on different AI2-THOR floorplans not included in the dataset. Consequently, all the tasks in the dataset are considered unseen during testing. We evaluate SMART-LLM with various language models as its backbone. We employ GPT-4 \cite{openai2023gpt4}, GPT-3.5 \cite{brown2020language}, Llama-2-70B \cite{touvron2023llama}, and Claude-3-Opus \cite{anthropic} to assess SMART-LLM's performance across diverse tasks and with various language models. We also compare our method to two alternative baselines. In the first baseline, we use our task decomposition method and prompts and randomly assign sub-tasks to available robots. The second baseline uses our task decomposition method along with a rule-based method for task allocation implemented based on \cite{gunn2015dynamic}.  Both baseline methods utilize GPT-4 to perform the task decomposition.

\subsection{Real-Robot Experiments}
In our real experiments with mobile robots, we assess the efficacy of SMART-LLM in handling tasks such as addressing visibility coverage challenges \cite{alam2020stochastic} and capturing images of objects. These tasks encompass diverse regions of varying sizes that necessitate visibility coverage and objects requiring image capture. Both aerial and ground robots, each with unique skill sets and visibility capabilities, are at our disposal for task execution. SMART-LLM is utilized to generate task plans according to these specific requirements. The number of robots required for achieving complete visibility coverage is contingent upon the size of the region and the capabilities of the robots involved. We presume that our robots are endowed with essential low-level skills, including \texttt{GoToLocation}, \texttt{ClickPicture}, and \texttt{Patrol}, essential for proficient task execution. To formulate task plans within this framework, we rely on the same prompt samples employed in our simulation experiments, which are grounded in the AI2-THOR simulator.

\subsection{Evaluation Metrics}
We employ five evaluation metrics: Success Rate (\textit{SR}), Task Completion Rate (\textit{TCR}), Goal Condition Recall (\textit{GCR}), Robot Utilization (\textit{RU}), and Executability (\textit{Exe}), following the methodology of \cite{singh2023progprompt}. Our evaluations are based on the dataset's final ground truth states, which we compare to the achieved states post-execution to assess task success.

\begin{itemize}
    \item \textit{Exe} is the fraction of actions in the task plan that can be executed, regardless of their impact on task completion.
    \item \textit{RU} evaluates the efficiency of the robot team by comparing the experiment's transition count to the dataset's ground truth transition count. \textit{RU} equals $1.0$ when they match, $0$ when transitions equal sub-task count, and falls between $0$ and $1$ otherwise.
    \item \textit{GCR} is quantified using the set difference between ground truth final state conditions and final state achieved, divided by the total number of task-specific goals in the dataset. 
    \item \textit{TCR} indicates task completion, irrespective of the robot utilization. If \textit{GCR} = $1$, then \textit{TCR} = $1$ else $0$.
    \item \textit{SR} is success rate and is $1$ when both $GCR$ and $RU$ are $1$, else it is $0$. The task is considered successful when completed with appropriate robot utilization.
\end{itemize}


\section{Results and Discussion}
\label{sec:results}

\subsection{Simulation Experiments}
Table \ref{tab:results} summarizes the average results across each category in the dataset for our method with various LLM backbones and baseline methods on unseen dataset tasks. Overall, SMART-LMM consistently delivers favorable outcomes irrespective of the LLM backbone employed. In the elemental task, SMART-LLM adeptly decomposed the given task and assigned the robot accordingly, except when employing the GPT-3.5 backbone, which encountered challenges in decomposing certain tasks. However, when accurate task decompositions were provided, the baseline method with random allocation performed successfully, given that that all robots possessed all the necessary skills.

In the simple tasks, the outcomes hinged on the LLM's capacity to decompose the given task in the appropriate sequence for execution. Notably, SMART-LLM utilizing Claude-3 as the backbone achieved superior results, although other LLMs also demonstrated commendable performance. GPT-4 and Claude-3 attain a perfect \textit{TCR} score of $1.0$ but have a lower \textit{SR} of $0.62$ and $0.87$ due to sequential execution instead of parallel execution by two robots, hence impacting \textit{RU}. Random task allocation often faltered, whereas rule-based allocation succeeded when task decompositions from LLM followed a logical sequence, yielding identical results to those achieved using LLM for allocation. 

In compound and complex tasks, our method consistently achieves favorable results across all LLM backbones, with a success rate of  $70\%$. We observed occasional struggles with task sequencing and robot team assignment in SMART-LLM, which may be mitigated by including additional prompt samples. However, the token limitations of certain LLMs hinder this optimization. Particularly, GPT-3.5 demonstrates underperformance compared to other LLM models, likely due to its deficiency in logical reasoning capabilities. Interestingly, Lllam 2, with only 70B parameters compared to trillions in other models, performs equally well. This success can be attributed to the prompting structure of SMART-LMM, enabling efficient performance even with smaller and simpler models. Consequently, SMART-LLM is deployable on local machines as well. Our decomposition method, employing random allocation, generally falters for skill-based task assignments due to its inability to consider the environment's state and the robot's skills. Rule-based allocation demonstrates satisfactory performance for compound tasks requiring the identification of robots with the appropriate skills. However, it falters in compound tasks involving object properties and complex tasks where team formation relies on specialized constraints. While these shortcomings could be mitigated by incorporating additional constraints into the code, this approach would require continual modifications or additions to accommodate new scenarios. Such practices compromise the scalability and ease of adaptation of the method. This underscores the scalability of SMART-LLM, as it does not necessitate any modifications for newer tasks, rendering the method highly scalable. Videos showing all the experiments can be accessed via \href{https://youtu.be/TnyCKwgTm3U}{https://youtu.be/TnyCKwgTm3U}.

\noindent\textbf{Infeasible Scenarios.~} In addition to the results presented in Table \ref{tab:results}, we conducted assessments involving more intricate tasks for which none of the robots possessed the required skills. This particular scenario is not included in Table \ref{tab:results} because no feasible code can be generated for the metrics to be measured. Notably, our approach utilizing the GPT-4 and Claude-3 backbones exhibited the capacity to discern this situation and refrained from generating any task allocation plan. In contrast, our method employing GPT-3.5 and Llama2 produced a task allocation plan involving robots ill-suited for the designated tasks. This disparity underscores the enhanced logical reasoning capabilities of GPT-4 and Claude-3 in recognizing and responding to such scenarios.

\noindent\textbf{Variability in Performance.~} The inherent non-deterministic characteristics of LLM introduce a degree of variability in its outcomes \cite{ouyang2023llm}. To assess this variability, we conducted 5 separate runs, each on a randomly selected task from every category within our dataset. Table \ref{tab:perf} provides the mean and standard deviations of the results observed across these trials for our approach using GPT-4 as the backbone. For elemental, simple, and complex tasks, our method consistently yielded comparable results. Nevertheless, in the case of complex scenarios, we encountered inconsistency, leading to occasional failures in robot task allocation.

\begin{table} [h]  
\centering
\scriptsize{}
\large
\caption{Variability in performance.}
\label{tab:perf}
\resizebox{\columnwidth}{!}{
\begin{tblr}{
  cells = {c},
  hline{1-2,6} = {-}{},
}
\textbf{Method}       & \textbf{SR} &  & \textbf{TCR} &  & \textbf{GCR} &  & \textbf{RU} &  & \textbf{Exe} &  \\
Elemental         & 1.00$\pm$0.00  &  & 1.00$\pm$0.00   &  & 1.00$\pm$0.00   &  & 1.00$\pm$0.00  &  & 1.00$\pm$0.00  &  \\
Simple  & 1.00$\pm$0.00  &  & 1.00$\pm$0.00   &  & 1.00$\pm$0.00   &  & 1.00$\pm$0.00  &  & 1.00$\pm$0.00  &  \\
Compound   & 1.00$\pm$0.00  &  & 1.00$\pm$0.00   &  & 1.00$\pm$0.00   &  & 1.00$\pm$0.00  &  & 1.00$\pm$0.00  &  \\
Complex      & 0.48$\pm$0.40  &  & 0.48$\pm$0.40   &  & 0.73$\pm$0.22   &  & 1.00$\pm$0.00  &  & 0.81$\pm$0.15  &  
\end{tblr}}
\vspace{-3mm}
\end{table}

\noindent\textbf{Ablation Study.~}
We utilized a benchmark dataset to evaluate different variations of our method, examining the impact of comments (both line-by-line and task summaries) in Python prompts. We validated our method with prompts lacking such comments. Additionally, we studied the influence of the coalition formation stage by removing it and directly allocating tasks based on task decomposition output. Table \ref{tab:abs} summarizes the ablations of our method using GPT-4 as the backbone. Removing comments generally reduces the success rate, underlining the value of natural language instructions with code. Notably, when comments are removed, task decomposition and allocation perform similarly across simple and elemental tasks but suffer in compound and complex tasks, indicating that comments aid in understanding reasoning and logical structures. The removal of coalition formation led to a decrease in the success rate. This decline was primarily attributed to the absence of detailed rational reasoning for task allocation. Without coalition formation, elemental tasks deviated the most, and the success rate dropped from $1.0$ to $0.66$, as all task allocation samples involved scenarios requiring robot teaming, leading to unnecessary multi-robot allocation for elemental tasks.

\begin{table}[t]
\centering

\scriptsize{}
\caption{Ablation studies.}
\label{tab:abs}
\resizebox{\columnwidth}{!}{
\begin{tblr}{
  cells = {c},
  hline{1-2,7} = {-}{},
}
\textbf{Method}       & \textbf{SR} &  & \textbf{TCR} &  & \textbf{GCR} &  & \textbf{RU} &  & \textbf{Exe} &  \\
Ours        & 0.75  &  & 0.90   &  & 0.94   &  & 0.88  &  & 0.99  &  \\
No Comments  & 0.48  &  & 0.65   &  & 0.73   &  & 0.75  &  & 0.78  &  \\
No Summary   & 0.61  &  & 0.74   &  & 0.80   &  & 0.78  &  & 0.81  &  \\
No Comm. \& Summ.      & 0.41  &  & 0.61   &  & 0.66   &  & 0.59  &  & 0.69  &  \\
No Coalition      & 0.60  &  & 0.68   &  & 0.75   &  & 0.85  &  & 0.82  &  \\
\end{tblr}
}
\vspace{-7mm}
\end{table}

\subsection{Real-Robot Experiments}
\begin{figure}[h]
    \centering
    \includegraphics[width=0.92\linewidth]{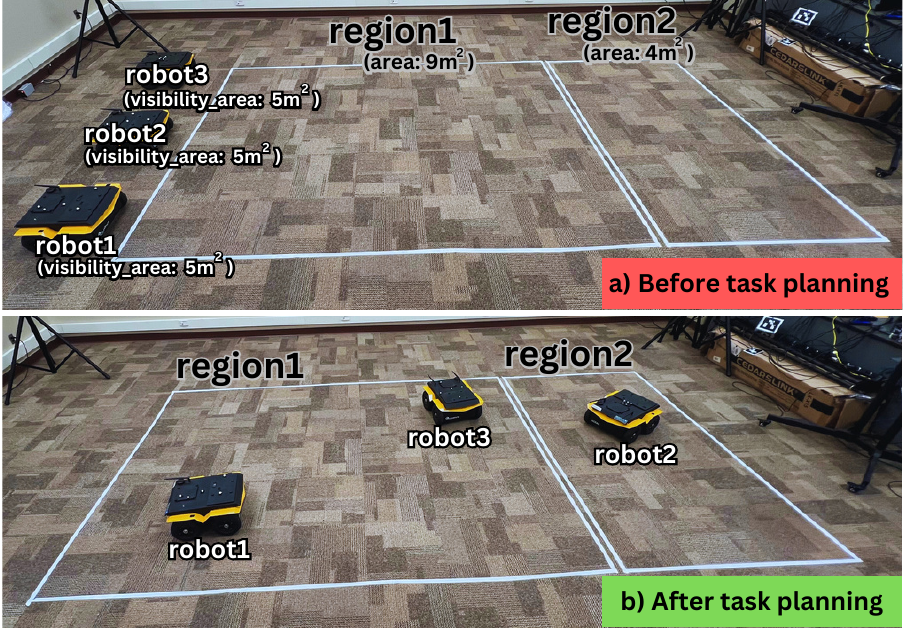}
    \caption{\textbf{Real-robot experiment}: a) team of robots and the regions to be patrolled; b) robots after task planning and patrolling their respective regions allocated based on visibility area.}   
    \label{fig:real}
    \vspace{-6mm}
\end{figure}

In real-robot experiments, we first tested our method for coverage visibility tasks with regions of different areas and robots with different visibility areas. When tested across various tasks, our method correctly generated task plans and allocated an appropriate number of robots. Despite this task being completely unseen and there were no sample prompts involving properties such as visibility, our method executed it seamlessly using real robots, bridging the gap between simulation and real-world applications. In Fig. \ref{fig:real}, for the instruction \textit{``patrol the regions"}, one or more robots are assigned to regions based on their visibility, and they patrol those regions. Furthermore, we evaluated our approach in tasks involving navigation and capturing images of predetermined objects. Despite these skills being entirely unseen, SMART-LLM successfully generated plans in the correct sequence and captured images of the specified objects.


\section{Conclusions and Future Work}
\label{sec:conclusion}
In our research, we delve into the potential of LLMs in the realm of generating task plans for heterogeneous robot teams. Our approach introduces prompting techniques, tailored to enhance the efficiency of the four key stages of multi-robot task planning. Each prompt takes into account the attributes of the environment and the capabilities of the individual robots, to generate a task plan. 

Our experiments validate that the proposed method can handle task instructions of varying complexities. Notably, our approach exhibits remarkable adaptability, allowing it to seamlessly generalize to new and unexplored environments, robot types, and task scenarios. This method streamlines the transition from simulations to real-world robot applications, enabling task plan samples from simulations to be used for generating task plans for real robot systems. In the future, we aim to enhance our work by implementing dynamic task allocation among robots and exploring multi-agent LLM frameworks for task planning.

\newpage
\bibliographystyle{IEEEtran}
\bibliography{references}

\begin{thebibliography}{10}
\providecommand{\url}[1]{#1}
\csname url@rmstyle\endcsname
\providecommand{\newblock}{\relax}
\providecommand{\bibinfo}[2]{#2}
\providecommand\BIBentrySTDinterwordspacing{\spaceskip=0pt\relax}
\providecommand\BIBentryALTinterwordstretchfactor{4}
\providecommand\BIBentryALTinterwordspacing{\spaceskip=\fontdimen2\font plus
\BIBentryALTinterwordstretchfactor\fontdimen3\font minus \fontdimen4\font\relax}
\providecommand\BIBforeignlanguage[2]{{%
\expandafter\ifx\csname l@#1\endcsname\relax
\typeout{** WARNING: IEEEtran.bst: No hyphenation pattern has been}%
\typeout{** loaded for the language `#1'. Using the pattern for}%
\typeout{** the default language instead.}%
\else
\language=\csname l@#1\endcsname
\fi
#2}}

\bibitem{benavidez2015design}
P.~Benavidez, M.~Kumar, S.~Agaian, and M.~Jamshidi, ``{Design of a Home Multi-robot System for the Elderly and Disabled},'' in \emph{2015 10th System of Systems Engineering Conference (SoSE)}, 2015.

\bibitem{queralta2020collaborative}
J.~P. Queralta, J.~Taipalmaa, B.~C. Pullinen, V.~K. Sarker, T.~N. Gia, H.~Tenhunen, M.~Gabbouj, J.~Raitoharju, and T.~Westerlund, ``{Collaborative Multi-Robot Search and Rescue: Planning, Coordination, Perception, and Active Vision},'' \emph{IEEE Access}, 2020.

\bibitem{chen2021integrated}
Z.~Chen, J.~Alonso-Mora, X.~Bai, D.~D. Harabor, and P.~J. Stuckey, ``{Integrated Task Assignment and Path Planning for Capacitated Multi-Agent Pickup and Delivery},'' \emph{IEEE Robotics and Automation Letters}, 2021.

\bibitem{rizk2019cooperative}
Y.~Rizk, M.~Awad, and E.~W. Tunstel, ``{Cooperative Heterogeneous Multi-Robot Systems: A Survey},'' \emph{ACM Computing Surveys (CSUR)}, 2019.

\bibitem{openai2023gpt4}
OpenAI, ``{GPT-4 Technical Report},'' 2023.

\bibitem{brown2020language}
T.~Brown, B.~Mann, N.~Ryder, M.~Subbiah, J.~D. Kaplan, P.~Dhariwal, A.~Neelakantan, P.~Shyam, G.~Sastry, A.~Askell, \emph{et~al.}, ``{Language Models are Few-shot Learners},'' \emph{Advances in neural information processing systems}, 2020.

\bibitem{touvron2023llama}
H.~Touvron, L.~Martin, K.~Stone, P.~Albert, A.~Almahairi, Y.~Babaei, N.~Bashlykov, S.~Batra, P.~Bhargava, S.~Bhosale, \emph{et~al.}, ``{Llama 2: Open Foundation and Fine-Tuned Chat Models},'' \emph{arXiv preprint arXiv:2307.09288}, 2023.

\bibitem{kolve2017ai2}
E.~Kolve, R.~Mottaghi, W.~Han, E.~VanderBilt, L.~Weihs, A.~Herrasti, M.~Deitke, K.~Ehsani, D.~Gordon, Y.~Zhu, \emph{et~al.}, ``{AI2-THOR: An Interactive 3D Environment for Visual AI},'' \emph{arXiv preprint arXiv:1712.05474}, 2017.

\bibitem{motes2020multi}
J.~Motes, R.~Sandstr{\"o}m, H.~Lee, S.~Thomas, and N.~M. Amato, ``{Multi-Robot Task and Motion Planning with Subtask Dependencies},'' \emph{IEEE Robotics and Automation Letters}, 2020.

\bibitem{shiarlis2018taco}
K.~Shiarlis, M.~Wulfmeier, S.~Salter, S.~Whiteson, and I.~Posner, ``{TACO: Learning Task Decomposition via Temporal Alignment for Control},'' in \emph{International Conference on Machine Learning}, 2018.

\bibitem{jansen2020visually}
P.~A. Jansen, ``{Visually-Grounded Planning without Vision: Language Models Infer Detailed Plans from High-Level Instructions},'' \emph{arXiv preprint arXiv:2009.14259}, 2020.

\bibitem{sakaguchi2021proscript}
K.~Sakaguchi, C.~Bhagavatula, R.~L. Bras, N.~Tandon, P.~Clark, and Y.~Choi, ``{prosScript: Partially Ordered Scripts Generation via Pre-trained Language Models},'' \emph{arXiv preprint arXiv:2104.08251}, 2021.

\bibitem{kong2015negotiation}
Y.~Kong, M.~Zhang, and D.~Ye, ``{A Negotiation-based Method for Task Allocation with Time Constraints in Open Grid Environments},'' \emph{Concurrency and Computation: Practice and Experience}, 2015.

\bibitem{braquet2021greedy}
M.~Braquet and E.~Bakolas, ``{Greedy Decentralized Auction-based Task Allocation for Multi-Agent Systems},'' \emph{IFAC-PapersOnLine}, 2021.

\bibitem{zitouni2020distributed}
F.~Zitouni, S.~Harous, and R.~Maamri, ``{A Distributed Approach to the Multi-Robot Task Allocation Problem using the Consensus-based Bundle Algorithm and Ant Colony System},'' \emph{IEEE Access}, 2020.

\bibitem{qin2021multi}
W.~Qin, Y.-N. Sun, Z.-L. Zhuang, Z.-Y. Lu, and Y.-M. Zhou, ``Multi-agent reinforcement learning-based dynamic task assignment for vehicles in urban transportation system,'' \emph{International Journal of Production Economics}, 2021.

\bibitem{barrett2015cooperating}
S.~Barrett and P.~Stone, ``{Cooperating with Unknown Teammates in Complex Domains: A Robot Soccer Case Study of Ad Hoc Teamwork},'' in \emph{Proceedings of the AAAI Conference on Artificial Intelligence}, 2015.

\bibitem{stegagno2013relative}
P.~Stegagno, M.~Cognetti, L.~Rosa, P.~Peliti, and G.~Oriolo, ``{Relative Localization and Identification in a Heterogeneous Multi-Robot System},'' in \emph{IEEE International Conference on Robotics and Automation}, 2013.

\bibitem{das2015distributed}
G.~P. Das, T.~M. McGinnity, S.~A. Coleman, and L.~Behera, ``{A Distributed Task Allocation Algorithm for a Multi-Robot System in Healthcare Facilities},'' \emph{Journal of Intelligent \& Robotic Systems}, 2015.

\bibitem{mina2020adaptive}
T.~Mina, S.~S. Kannan, W.~Jo, and B.-C. Min, ``{Adaptive Workload Allocation for Multi-Human Multi-Robot Teams for Independent and Homogeneous Tasks},'' \emph{IEEE Access}, 2020.

\bibitem{liu2016coalition}
Z.~Liu, X.-g. Gao, and X.-w. Fu, ``{Coalition Formation for Multiple Heterogeneous UAVs Cooperative Search and Prosecute with Communication Constraints},'' in \emph{Chinese Control and Decision Conference (CCDC)}, 2016.

\bibitem{jones2006dynamically}
E.~G. Jones, B.~Browning, M.~B. Dias, B.~Argall, M.~Veloso, and A.~Stentz, ``{Dynamically Formed Heterogeneous Robot Teams Performing Tightly-Coordinated Tasks},'' in \emph{IEEE International Conference on Robotics and Automation}, 2006.

\bibitem{padmanabhan2015coalition}
M.~Padmanabhan and G.~Suresh, ``{Coalition Formation and Task Allocation of Multiple Autonomous Robots},'' in \emph{2015 3rd International Conference on Signal Processing, Communication and Networking (ICSCN)}, 2015.

\bibitem{liu2022embodied}
X.~Liu, X.~Li, D.~Guo, S.~Tan, H.~Liu, and F.~Sun, ``{Embodied multi-agent task planning from ambiguous instruction},'' \emph{Robotics: Science and Systems}, 2022.

\bibitem{madaan2022language}
A.~Madaan, S.~Zhou, U.~Alon, Y.~Yang, and G.~Neubig, ``{Language Models of Code are Few-shot Commonsense Learners},'' \emph{arXiv preprint arXiv:2210.07128}, 2022.

\bibitem{huang2022language}
W.~Huang, P.~Abbeel, D.~Pathak, and I.~Mordatch, ``{Language Models as Zero-Shot Planners: Extracting Actionable Knowledge for Embodied Agents},'' in \emph{International Conference on Machine Learning}, 2022.

\bibitem{ahn2022can}
M.~Ahn, A.~Brohan, N.~Brown, Y.~Chebotar, O.~Cortes, B.~David, C.~Finn, C.~Fu, K.~Gopalakrishnan, K.~Hausman, \emph{et~al.}, ``{Do As I Can, Not As I Say: Grounding Language in Robotic Affordances},'' \emph{arXiv preprint arXiv:2204.01691}, 2022.

\bibitem{lin2023text2motion}
K.~Lin, C.~Agia, T.~Migimatsu, M.~Pavone, and J.~Bohg, ``{Text2Motion: From Natural Language Instructions to Feasible Plans},'' \emph{arXiv preprint arXiv:2303.12153}, 2023.

\bibitem{singh2023progprompt}
I.~Singh, V.~Blukis, A.~Mousavian, A.~Goyal, D.~Xu, J.~Tremblay, D.~Fox, J.~Thomason, and A.~Garg, ``{ProgPrompt: Generating Situated Robot Task Plans using Large Language Models},'' in \emph{2023 IEEE International Conference on Robotics and Automation}, 2023.

\bibitem{chen2023open}
B.~Chen, F.~Xia, B.~Ichter, K.~Rao, K.~Gopalakrishnan, M.~S. Ryoo, A.~Stone, and D.~Kappler, ``{Open-Vocabulary Queryable Scene Representations for Real World Planning},'' in \emph{2023 IEEE International Conference on Robotics and Automation}, 2023.

\bibitem{wu2023tidybot}
J.~Wu, R.~Antonova, A.~Kan, M.~Lepert, A.~Zeng, S.~Song, J.~Bohg, S.~Rusinkiewicz, and T.~Funkhouser, ``{TidyBot: Personalized Robot Assistance with Large Language Models},'' \emph{Autonomous Robots}, 2023.

\bibitem{huang2023instruct2act}
S.~Huang, Z.~Jiang, H.~Dong, Y.~Qiao, P.~Gao, and H.~Li, ``{Instruct2Act: Mapping Multi-modality Instructions to Robotic Actions with Large Language Model},'' \emph{arXiv preprint arXiv:2305.11176}, 2023.

\bibitem{vemprala2023chatgpt}
S.~Vemprala, R.~Bonatti, A.~Bucker, and A.~Kapoor, ``{ChatGPT for Robotics: Design Principles and Model Abilities},'' \emph{Microsoft Auton. Syst. Robot. Res}, 2023.

\bibitem{huang2022inner}
W.~Huang, F.~Xia, T.~Xiao, H.~Chan, J.~Liang, P.~Florence, A.~Zeng, J.~Tompson, I.~Mordatch, Y.~Chebotar, \emph{et~al.}, ``{Inner Monologue: Embodied Reasoning through Planning with Language Models},'' \emph{arXiv preprint arXiv:2207.05608}, 2022.

\bibitem{yao2022react}
S.~Yao, J.~Zhao, D.~Yu, N.~Du, I.~Shafran, K.~Narasimhan, and Y.~Cao, ``{React: Synergizing Reasoning and Acting in Language Models},'' \emph{arXiv preprint arXiv:2210.03629}, 2022.

\bibitem{talebirad2023multi}
Y.~Talebirad and A.~Nadiri, ``{Multi-Agent Collaboration: Harnessing the Power of Intelligent LLM Agents},'' \emph{arXiv preprint arXiv:2306.03314}, 2023.

\bibitem{liu2023bolaa}
Z.~Liu, W.~Yao, J.~Zhang, L.~Xue, S.~Heinecke, R.~Murthy, Y.~Feng, Z.~Chen, J.~C. Niebles, D.~Arpit, \emph{et~al.}, ``{BOLAA: Benchmarking and Orchestrating LLM-Augmented Autonomous Agents},'' \emph{arXiv preprint arXiv:2308.05960}, 2023.

\bibitem{hong2023metagpt}
S.~Hong, X.~Zheng, J.~Chen, Y.~Cheng, C.~Zhang, Z.~Wang, S.~K.~S. Yau, Z.~Lin, L.~Zhou, C.~Ran, \emph{et~al.}, ``{MetaGPT: Meta Programming for Multi-Agent Collaborative Framework},'' \emph{arXiv preprint arXiv:2308.00352}, 2023.

\bibitem{wu2023reasoning}
Z.~Wu, L.~Qiu, A.~Ross, E.~Aky{\"u}rek, B.~Chen, B.~Wang, N.~Kim, J.~Andreas, and Y.~Kim, ``{Reasoning or Reciting? Exploring the Capabilities and Limitations of Language Models through Counterfactual Tasks},'' \emph{arXiv preprint arXiv:2307.02477}, 2023.

\bibitem{gunn2015dynamic}
T.~Gunn and J.~Anderson, ``Dynamic heterogeneous team formation for robotic urban search and rescue,'' \emph{Journal of Computer and System Sciences}, 2015.

\bibitem{anthropic}
\BIBentryALTinterwordspacing
``{Anthropic Claude 3}.'' [Online]. Available: \url{https://www.anthropic.com/news/claude-3-family}
\BIBentrySTDinterwordspacing

\bibitem{alam2020stochastic}
T.~Alam, M.~M. Rahman, P.~Carrillo, L.~Bobadilla, and B.~Rapp, ``{Stochastic Multi-Robot Patrolling with Limited Visibility},'' \emph{Journal of Intelligent \& Robotic Systems}, 2020.

\bibitem{ouyang2023llm}
S.~Ouyang, J.~M. Zhang, M.~Harman, and M.~Wang, ``{LLM is Like a Box of Chocolates: The Non-determinism of ChatGPT in Code Generation},'' \emph{arXiv preprint arXiv:2308.02828}, 2023.

\end{thebibliography}
\end{document}